\pdfoutput=1

\documentclass[11pt]{article}

\usepackage[]{ACL2023}

\usepackage{times}
\usepackage{latexsym}

\usepackage[T1]{fontenc}

\usepackage[utf8]{inputenc}

\usepackage{microtype}

\usepackage{inconsolata}
\usepackage{subfigure}
\usepackage{graphicx}
\usepackage{amsmath}
\usepackage[final]{changes}

\title{Increasing Diversity While Maintaining Accuracy: Text Data Generation with Large Language Models and Human Interventions}

\author{John Joon Young Chung \\
  University of Michigan \\
  \texttt{jjyc@umich.edu} \\\And
  Ece Kamar \\
  Microsoft Research \\
  \texttt{eckamar@microsoft.com} \\\And
  Saleema Amershi \\
  Microsoft Research \\
  \texttt{samershi@microsoft.com}
  }

\begin{document}
\maketitle
\begin{abstract}
Large language models (LLMs) can be used to generate text data for training and evaluating other models. However, creating high-quality datasets with LLMs can be challenging. In this work, we explore human-AI partnerships to facilitate high diversity and accuracy in LLM-based text data generation. We first examine two approaches to diversify text generation: 1) logit suppression, which minimizes the generation of languages that have already been frequently generated, and 2) temperature sampling, which flattens the token sampling probability. We found that diversification approaches can increase data diversity but often at the cost of data accuracy (i.e., text and labels being appropriate for the target domain). To address this issue, we examined two human interventions, 1) label replacement (LR), correcting misaligned labels, and 2) out-of-scope filtering (OOSF), removing instances that are out of the user's domain of interest or to which no considered label applies. With oracle studies, we found that LR increases the absolute accuracy of models trained with diversified datasets by 14.4\%. Moreover, we found that some models trained with data generated with LR interventions outperformed LLM-based few-shot classification. In contrast, OOSF was not effective in increasing model accuracy, implying the need for future work in human-in-the-loop text data generation.
\end{abstract}

\section{Introduction}
Training custom natural language classification models has become easier with many tools (e.g., Huggingface\footnote{https://huggingface.co/}).
However, data collection remains a costly part of model building. For example, existing open-source datasets may not be usable if they do not match the distribution of a model builder's target domain or do not contain desired labels. In such cases, the model builder may need to collect and label new data which could be costly (e.g., in terms of the time and resources to scrape data or pay people to generate or annotate new data). 

Advances in generative large language models (LLMs), such as GPT-3~\cite{brown2020language}, present a novel approach for creating training data for classification models~\cite{yoo2021gpt3mix, sahu2022data, kumar2020data}. 
Model builders can prompt an LLM with the domain of texts and labels of interest and the LLM can quickly generate text data for the model builder’s needs.
This approach allows model builders to acquire a large amount of data even when they initially have no or few data instances. 
With the generated data, the model builder can train a separate affordable model (e.g., BERT~\cite{devlin2019bert}) to perform the specific task.

While LLMs can directly support this classification task with few-shot learning, it might not be the best option for every model builder---some might not have enough resources (e.g., GPUs) or budget (e.g., credit for GPT-3) to run expensive models. Others might be concerned about privacy or security issues when they use LLMs from external APIs (e.g., OpenAI API). In such cases, generating data from LLMs and training custom models could be a more viable approach. Moreover, if we share generated datasets within the community, we can also benefit those who do not have access to LLMs. Lastly, we can also use generated datasets to test models. With these benefits of generating new text datasets with LLMs, the practical concern is how to generate high-quality datasets.

In this work, we investigate human-AI partnerships to efficiently create high-quality datasets with LLM-based text generation. High-quality datasets should have high diversity and coverage, informing the extent of data that the model may encounter. At the same time, the generated text should have high accuracy, being relevant to the model’s target task while having accurate accompanying labels. To these ends, we first study two technical approaches to diversify text generation (Section~\ref{sec:div}): 1) logit suppression, which diversifies the generated texts by decreasing the probability of sampling tokens that have already appeared frequently in the previous generation, and 2) temperature sampling, which flattens the probability distribution of sampled tokens to pick less likely texts. From an experiment on eight classification tasks with GPT-3 as a text generator\added{ (Section~\ref{sec:exp1})}, we found that diversification approaches can have mixed results. While increasing data diversity, these approaches can hurt accuracy in generation and similarity to the original datasets for the task. 

We demonstrate that human interventions (Section~\ref{sec:human}) are the key to resolving these issues in text generation diversification. We examine human interventions of replacing inaccurate labels with accurate ones (label replacement) and filtering out-of-scope data (out-of-scope data filtering). With oracle studies\added{ (Section~\ref{sec:exp2})}, we found that replacing all incorrect labels increased model accuracy by 14.4\% when we used both logit suppression and high temperature. This performance increase brings in practical benefits---without label replacement, the average accuracy of models trained with GPT-3-generated data was lower than that of GPT-3 classification with few-shot learning, but with 180 instances label-replaced, the models trained with generated data started to outperform GPT-3 few-shot classification. Out-of-scope data filtering had limited utility in increasing model accuracy, possibly due to the negative impact of removing training instances. We discuss how human interventions can further facilitate the diversity and accuracy of text data generation.  

Our contributions are: 
\begin{itemize}
    \item A methodolgy that combines LLM generation approaches and human supervision for diversified and accurate data generation. 
    \item An experiment showing how text generation diversification impacts the accuracy of trained models and other qualities of the data, such as diversity and accuracy in the generation. 
    \item Oracle studies on how human effort to replace misaligned labels and filter out-of-scope data instances can impact the performance of models trained on data generated with text diversification. 
\end{itemize}

\section{Related Work}

\subsection{Text Data Generation for Model Training}

In NLP, data augmentation, where data are multiplied based on existing data, is one context where text data are generated for model training.
There were many approaches, from replacing words with synonyms~\cite{wei2019eda, zhang2015character}, to randomly editing texts~\cite{wei2019eda}, predicting replaceable words~\cite{ng2020ssmba}, back-translating~\cite{fadaee2017data}, generating label-flipped data~\cite{zhou2022flipda}, or using reinforcement learning to condition generation~\cite{liu2020data}.
Inspired by MixUp~\cite{zhang2018mixup}, which mixes different examples in vision data, researchers also blended texts to augment data~\cite{guo2020sequence, sun2020mixup, zhang2022treemix}.
Other approaches generate texts by learning from different datasets~\cite{xia2020cgbert, hou2018sequence, chen2020mixtext, yoo2019data}.

Recently, with the generative capacity of LLMs, researchers proposed generating datasets with zero or very few samples and training a separate model to serve the specific task~\cite{kumar2020data, yoo2021gpt3mix, sahu2022data, yuan2021synthbio, hartvigsen2022toxigen}. 
As this approach would extract information from large models, they would be analogous to knowledge distillation~\cite{phuong2019towards, hinton2014distilling} or dataset distillation~\cite{wang2018dataset, cazenavette2022distillation}. 
LLM-generated data has also been used to test other trained models~\cite{ribeiro2022adaptive, perez2022red}. 
In this work, we extend the previous work by investigating the generation of high-quality data with accurate diversification. 

\subsection{Text Generation with LLMs}
As the size of language models increases, researchers found that LLMs can serve different generation tasks based on input prompts and examples~\cite{brown2020language}. 
This approach can be used to generate text data with instructional prompts and a few examples.
However, for the generated data to be useful, diversity and coverage should be ensured.
Control of the sampling temperature~\cite{goodfellow2016deep} would be relevant, as it facilitates the unlikely generation, but it was not evaluated for the facilitation of diversity and coverage. 
Inspired by previous work on controlling LLM generation, we examine human-AI approaches to steer data generation to have higher diversity while securing accuracy in the alignment of specified labels. 

\subsection{Human-In-The-Loop}

Human interventions are imperative to train high-performance machine learning models, as people curate datasets, configure model architectures, and test the trained models. 
Researchers investigated approaches to make human interventions more interactive in model training pipelines, by closing gaps between model training and data curation~\cite{fogarty2008cueflik, amershi2009overview, amershi2012regroup, levonian2022tradeoffs}, humans extracting features~\cite{branson2010visual, cheng2015flock}, interactively changing the error patterns~\cite{kapoor2010interactive, talbot2009ensemblematrix}, or interactively testing models~\cite{wu2019errudite, yan2022isea, ribeiro2020beyond, cabrera2021discovering, suh2019anchorviz}. Generative models introduce novel approaches to interactively tune and evaluate models by leveraging generated results as data instances for training and testing~\cite{ribeiro2022adaptive}. In this work, we explored harnessing diversified and accurate datasets by combining LLM-based text generation and human interventions.
\section{Diversified Text Data Generation}
\label{sec:div}
We lay out the desired characteristics of the datasets for model building. Then, we introduce approaches to generate diversified datasets with LLMs.  

\subsection{Goals}

Ideal classification datasets need to have the following characteristics: 
1) Scoped: fall in the model builder's domain of interest while classifiable with labels of interest,
2) Label accurate: accompany accurate labels,
and 3) Diverse: cover cases the model would encounter during test time.
These goals are difficult to achieve simultaneously but need to be balanced. Only considering diversity, randomly generating any text would be enough, but it would hurt scope and label accuracy. Likewise, only considering the scope and label accuracy, generating an accurate but limited variety of text would be enough, but it would hurt the diversity. 

\subsection{Diversifying Approaches}
We introduce the setting to use LLM-based data generation for model training. Then, we lay out two approaches to promote diversity in text data generation. We also note their potential risks of harming the scope and accuracy.

\subsubsection{Settings for Data Generation}

When prompting LLMs, we consider 1) a text type and 2) labels in the prompts. 
While there can be many different prompts, in our paper, we used the following prompt:
\begin{equation}\label{prompt_A}
\footnotesize
\parbox{\dimexpr\linewidth-3em}
{Write \textbf{a movie review (\texttt{text type})} to cover all following elements\\
Elements: \textbf{positive sentiment (\texttt{label})}\\
\textbf{Movie review (\texttt{text type})}: "This is a great movie"}
\tag{A}
\end{equation}

\normalsize
Model builders can also prepend examples in the same format. 
The generation process is iterative, and model builders can use intermediate data points as examples in later prompts. The model builders can generate data until they reach the desired number of data points. 
With the generated data, the model builder would finetune a separate smaller model that serves the target task. 
With this approach of finetuning a smaller model, there can be a question of whether finetuning a separate model would result in higher accuracy than using zero-shot or few-shot learning of the LLM. In the later study, we show the cases where finetuned smaller models perform better than the LLM.

\begin{figure}
    \centering
    \includegraphics[width=.478\textwidth]{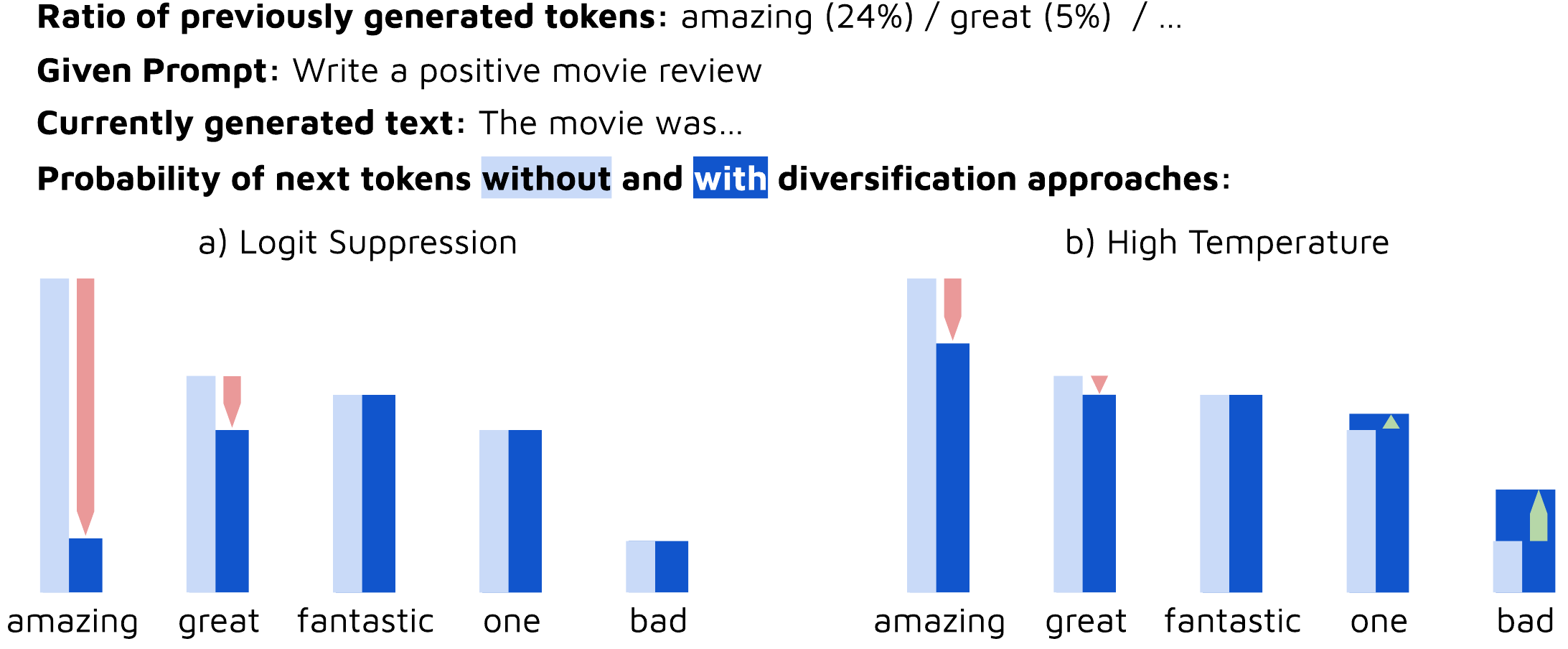}
    \caption{Examples of Diversification Approaches.}
    \label{fig:diversification}
\end{figure}

\subsubsection{Logit Suppression}

Logit suppression is a diversification approach that suppresses tokens that have already been generated frequently in the intermediate dataset (Figure~\ref{fig:diversification}a). 
With this approach, the generation pipeline logs the frequency of tokens that have been generated so far. Then, to diversify the selection of tokens, logit suppression decreases the probability of high-frequency tokens. However, with this approach, some tokens that could contribute to accurate generation can be suppressed. 

\subsubsection{High Temperature}

The temperature of sampling distribution~\cite{goodfellow2016deep} controls how ``flat'' the token sampling probability is (the equation is explained in Appendix~\ref{appendix:temp_samp}). 
High temperature leads to “flatter” token sampling probabilities (Figure~\ref{fig:diversification}b), increasing the probability of sampling ``less likely'' tokens and diversifying generation. Similar to logit suppression, extremely high temperatures can result in tokens irrelevant to the prompt, hurting accuracy in generation results.

\section{Experiment1: Diversified Text Data Generation}
\label{sec:exp1}
We evaluated how diversification approaches impact the diversity of the generated data and the accuracy of models trained with the dataset. 

\subsection{Experiment Settings}

\subsubsection{Tasks} 
We used tasks from eight datasets. \textbf{SST-2}~\cite{socher2013recursive} is a binary sentiment classification dataset from Rotten Tomatoes movie reviews. Clickbait classification dataset (\textbf{CB})~\cite{chakraborty2016stop} is news headlines labeled either clickbait or non-clickbait. \textbf{CARER}~\cite{saravia2018carer} is Twitter statements labeled with one of the six emotion categories. \textbf{PubMed} 200k RCT~\cite{dernoncourt2017pubmed} has five classes regarding the roles of sentences in medical papers. The subjectivity dataset (\textbf{SUBJ}) is movie review texts labeled subjective or objective~\cite{pang2004sentimental}. Formality classification dataset (\textbf{FO})~\cite{squinky2015lahiri} has labels on whether the text is formal or informal. \textbf{HWU64}~\cite{liu2021benchmarking} is a dataset with human utterances to chatbots, and we used 18 domain classes for our experiments. Corpus of Linguistic Acceptability (\textbf{COLA})~\cite{warstadt2019neural} is publication texts with annotations on whether the text is grammatically correct or not.  

\subsubsection{Generation Method}

As a generative LLM, we used the \texttt{text-davinci-002} model of GPT-3 through OpenAI API Access with Prompt~\ref{prompt_A}.
We list the specific text types and labels used for each dataset in Appendix~\ref{sec:appendix_prompt_used}.  
The generation process was iterative, with 20 data points generated with a single prompt for each API call. As a single prompt can only generate data instances for a single label, the generation process cycled through all considered labels while balancing the number of instances for each class. 
As our tasks dealt with short text data, we limited the generation length to 100 tokens. We set the frequency penalty and top p to 0.02 and 1, respectively.
Except for SST-2, we generated 5600 instances for a single training dataset. For SST-2, we generated 6922 data points. We chose these numbers to ensure a low generation budget while having fair quality when training models.
Specifically, with a maximum length of 100 tokens for each generated instance, if the prompt includes examples for n classes, the number of required tokens for each instance would be (100+30) $\times$ (n+1) (where 30 come from the instructional prompts). 
With the generation pricing of \$0.02/1000 tokens for \texttt{text-davinci-002} model, 5600 and 6922 instances resulted in maximum spending of \$14.56 $\times$ (n+1) and \$17.80 $\times$ (n+1), respectively.  
In our pilot tests, model accuracy saturated after these numbers of instances. 
For the oracle training dataset, with which we compared the quality of the datasets, we sampled instances from the original training dataset for the task. The test dataset was sampled from the original test dataset. 
We provide details on how we sampled these instances in Appendix~\ref{appendix:oracle_dataset}.

\paragraph{Generation Conditions}
In addition to \textbf{logit suppression} and \textbf{temperature sampling}, we also consider \textbf{example seeding}, whether the generation pipeline begins with an initial set of example instances. 
We can use multiple approaches simultaneously (e.g., using logit suppression and temperature sampling together), and how these approaches interact is also the scope of our questions. For a single combination of conditions, we generated three datasets, as there could be some variance in the results with the initial seeds and the examples generated initially.   

\begin{figure*}
    \centering
    \includegraphics[width=\textwidth]{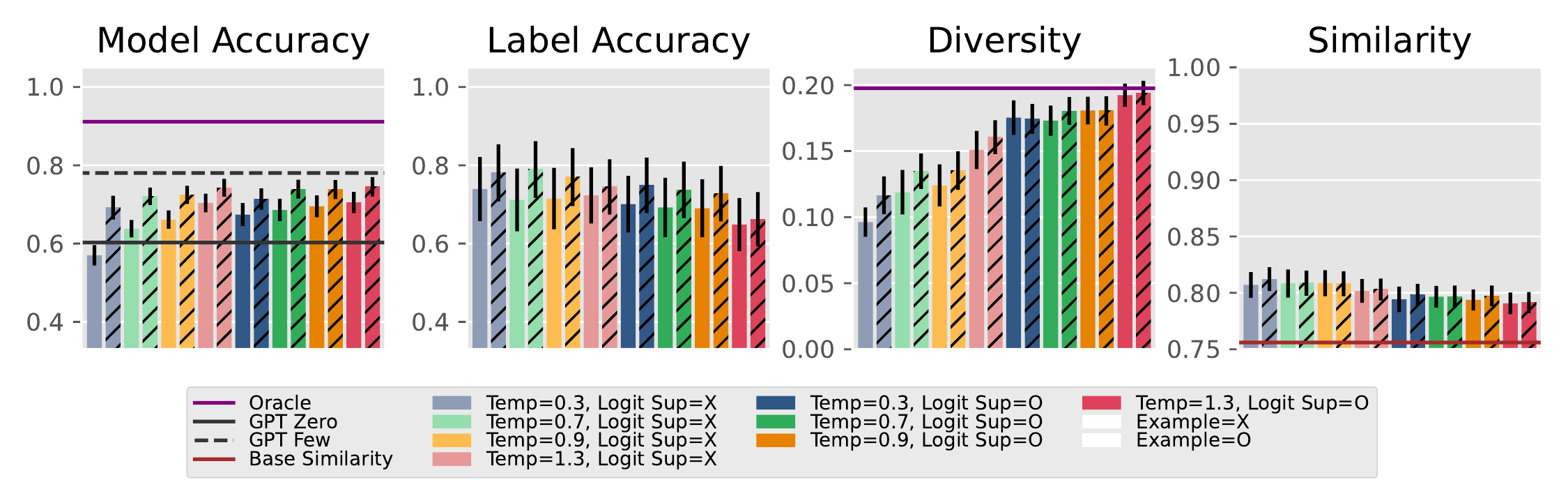}
    \caption{Impact of logit suppression and high temperatures on model accuracy, label accuracy, diversity, and similarity to the oracle dataset, averaged across eight tasks. Bars without hatches start generation without examples while those with hatches start with few-shot generation. Throughout this paper, error bars indicate 95\% confidence interval.}
    \label{fig:exp1_whole}
\end{figure*}

We instantiated \textbf{logit suppression} with the logit bias function in OpenAI API Access\footnote{\url{https://beta.openai.com/docs/api-reference/completions/create\#completions/create-logit_bias}}, which can increase or decrease the probability of sampling tokens. 
Every time we complete a single generation iteration, we recorded the frequency of tokens generated by GPT-3. 
As the OpenAI API only allows 100 tokens for logit biasing, we suppressed only the 100 most appeared tokens. Specifically, for the logit bias weights, we multiplied the token appearance ratio (in percentage) by -7.5 while capping the minimum weight at –7.5. 
For \textbf{temperature sampling}, we used four temperature values, 0.3, 0.7, 0.9, and 1.3. 
When \textbf{seeding examples}, we first randomly sampled 18 examples from oracle training data with a balanced number of labels. 
Only for PubMed, which has five classes, we used 15 seed examples. 
We used sampled data points as an initial example pool. 
With example seeding, from the first generation iteration, examples were randomly chosen from the pool. 
Without the seeding examples, we completed the first cycle of generations as a zero-shot generation.
After the first cycle, since we would have generated data instances for all labels, we added examples to the prompt. 
When adding examples, we randomly sampled the examples for all labels, one example for each label.

\subsubsection{Training Method}

With the generated data, we finetuned base size BERT~\cite{devlin2019bert} classifiers with 109M parameters using pretrained weights from the Huggingface Transformer library~\cite{wolf2020transformers} with a randomly initialized fully connected classifier layer. For each dataset, we trained the five different models with the same dataset. With three datasets for each combination of approaches, it resulted in 15 models for a condition. While training, Adam optimizer was used, with a learning rate of 3e-5 and a warm-up period of 3 epochs. We adopted the early stopping with the patience of five training epochs. We used PyTorch and RTX A6000 GPUs for training.

\subsection{Metrics}

We compared the accuracies of models trained with generated data to 1) models trained with oracle datasets (oracle model) and 2) GPT-3's few-/zero-shot classifications (\texttt{text-davinci-002}). 
For GPT-3 few-shot learning, we used 18 examples (15 only for PubMed) with the same number of examples for each label. 
We also measured the diversity of the dataset using Remote-Clique metric~\cite{cox2021directed}, which is the average mean pairwise distances. Specifically, we embedded generated data with BERT~\cite{devlin2019bert}, then calculated the distances. We also evaluated label accuracy, which is the accuracy of the alignment between the generated texts and the specified labels. For this metric, except for SST-2, we used the oracle model as the evaluator. For SST-2, we used GPT-3 few-shot classification as the evaluator, as it has higher accuracy than the oracle model. 
We also measured the similarity of the generated dataset to the oracle dataset with the average mean pairwise distances between the two. For similarity, we also used BERT to embed the generated texts.

\subsection{Results}

Figure~\ref{fig:exp1_whole} shows the results of the first experiment for all tasks. 
The first column shows the model accuracy results. It also shows the accuracy of zero-shot and few-shot GPT-3 classification (gray solid and dashed line, respectively) and the model trained with the oracle training dataset (purple line). 
The second column shows the label accuracy, and the third column shows the diversity. The diversity plots also show the diversity of oracle datasets (purple line). The last column shows the similarity. It also shows the base similarity (brown line), which is the average distance between all the different datasets that we considered. 

First, to evaluate how diversity, label accuracy, and similarity impact model accuracy, we performed a linear regression analysis. The analysis showed that label accuracy, diversity, and similarity are positively correlated with model accuracy, with significance (coef=.4797 and p<0.001 for label accuracy, coef=.2260 and p<0.001 for diversity, and coef=0.1980 and p<0.005 for similarity).

Regarding specific patterns, logit suppression increased diversity while hurting the label accuracy and the similarity to the oracle dataset. 
High temperature increased diversity and decreased label accuracy, but to a smaller degree than logit suppression. 
The application of each diversification approach increased the model accuracy, but when used together, the benefit did not add up.
For instance, in Model Accuracy of Figure~\ref{fig:exp1_whole}, each high temperature (1.3, red light bars) and logit suppression (dark blue bars) could increase the model accuracy from when using a low temperature (0.3, light blue bars). However, when using them together (dark red bars), the resulting accuracy was not much different from only using high temperatures (light red bars). It indicates that the effect of logit suppression has diminished by using high temperatures and logit suppression together.
Seeding examples increases label accuracy and model accuracy. Examples also slightly increased diversity when used without logit suppression.
Whether models trained with LLM-generated data would have higher accuracy than zero- or few-shot learning of LLMs depends on the task. We provide a detailed result on each task in Appendix~\ref{appendix:exp1_detail_results}.

\section{Human Interventions to Fix Inaccurate Text Generation}
\label{sec:human}
The first study shows that diversifying approaches can have mixed effects, hurting the accuracy in generation. 
We propose two human interventions to improve the generated data, based on issues that we found from qualitatively analyzing the generated data. 
The first is \textbf{label replacement (LR)}, switching the misaligned label to the correct one. 
The second is \textbf{out-of-scope data filtering (OOSF)}, which removes instances that are outside the domain of interest and do not match any labels (OOS instances). 

While LR and OOSF might facilitate accurate generation with diversifying approaches, inspecting all data points can require a lot of effort. 
Hence, we propose a simple way to scale the effort of the model builder, which is training a \textbf{proxy model}.
With this approach, model builders will first label a small number of data points. Then, with those labels, they will train binary classifiers as proxy models, where each learns about a single label (i.e., a label class from labels of interest or if the instance is out of scope). 
For unlabeled data points, proxy models can make inferences on behalf of the model builder. 
We introduced the specific implementation of this approach in Section~\ref{sec:exp2}.   

\section{Experiment2: Human Interventions For Diversifed Text Generation}
\label{sec:exp2}

We evaluated LR and OOSF. 
Except for adding LR and OOSF, we used the same tasks, datasets, training methods, and metrics as in Section~\ref{sec:exp1}. 
In this section, we focus on reporting results for two temperature values, 0.3 and 1.3. 
We present the results with the rest of the temperatures in Appendix~\ref{sec:appendix_temp}.
Also, in this section, when reporting, we merged conditions with and without example seeding.

\subsection{Experiment Settings}

\subsubsection{Label Replacement}

For LR, we conducted an oracle experiment. 
For each task, we used the highest accuracy model as the oracle labeler. 
Therefore, we used oracle models as a labeler, but only for SST-2, we used GPT-3 few-shot classification as a labeler. 
We conducted LR on the datasets generated in experiment 1. 

We had two approaches for LR: 1) do LR to all data points and 2) use proxy models with LR on partial data. 
For 1), we inspected all generated texts with simulated labelers and replaced labels as the labelers predicted. For 2), we sampled a set of instances from the generated dataset, applied the oracle labeler to them, and then trained proxy models with those data. Specifically, we sampled 90, 180, or 270 data instances. 
When training, for each class, we trained a proxy model that performs binary classification for the class.
For each proxy model, the data instances labeled with the target label were used as positive instances, while the rest were used as negative instances. 
We applied proxy models to the uninspected data to obtain confidence scores for each label. 
For each class, we calculated the final score as follows: 
\begin{equation}
    S_{f, i} = S_{s, i} * w + S_{p, i} * (1-w)
\end{equation}

where for the class $i$, $S_{f, i}$ is the final score, $S_{p, i}$ is the confidence score of the proxy model, $S_{s, i}$ is if the class is specified when generating the text (1 when the class is specified, 0 otherwise), and $w$ is the weighting constant. 
We considered $S_{s, i}$ as there can be a chance that the proxy model is inaccurate and the correct labels are swapped.
For our experiment, we used $w$ of 0.3. We chose the label with the highest final score as the label to be replaced. 
For training proxy models, we trained linear support vector classifiers with a maximum iteration of 10000 while using texts embedded with BERT~\cite{devlin2019bert} as input. 
We chose to train multiple proxy models for each class over training a single proxy model for all classes, as it tends to be more reliable in our pilots when there are many classes. As the labeling of the proxy model depends on the initial samples, for each generated dataset in experiment 1, we applied the approach five times. 

\subsubsection{Out-of-Scope Filtering}

\begin{table}[t]
\small
\centering
\begin{tabular}{ll|ll}
\hline
Task   & Ratio & Task   & Ratio \\ \hline
CARER  & 20.56\% & CB     & 1.39\%               \\
COLA   & 0.00\% & FO     & 0.56\%  \\
HWU64  & 0.28\% & PubMed & 1.11\%               \\
SST-2  & 3.61\% & SUBJ   & 3.06\%                \\\hline
\end{tabular}
\caption{Ratio of out-of-scope instances from 360 samples.}
\label{tab:OOS_ratio}
\end{table}

With OOSF, we first tried to understand how OOS instances occur. Therefore, we sampled 360 data instances for each task from the union of all the datasets generated for the task. 
Then, an author served as the oracle and annotated if they were OOS or not. 
Note that, as the definition of OOS instance, we filtered those instances that are outside the task domain or to which no label is applicable. 
We found that COLA, FO, HWU64, and PubMed have zero to four instances of OOS (Table~\ref{tab:OOS_ratio}). 
For the later analysis, we only considered the rest of the datasets, with at least five OOS instances. 
We present examples of OOS instances in Appendix~\ref{appendix:oos_instances}.

\begin{table}[t]
\small
\centering
\begin{tabular}{ll|ll}
\hline
Task   & Accuracy (std) & Task & Accuracy (std) \\ \hline
CARER  & 94.93 (2.20) & CB     & 100 (0.00)               \\
SST-2  & 97.18 (0.89) & SUBJ   & 97.5 (1.04)                 \\
\hline
\end{tabular}
\caption{OOSF proxy model performance. Note that CB only had five OOS instances, with one used for test.}
\label{tab:OOS_filter_model}
\end{table}

With the annotated data, we trained proxy models to annotate the instances unseen by the author, which were binary linear support vector classifiers with the maximum iteration of 10000 and BERT-embedded inputs. 
With the trained model, we did OOSF on the datasets generated in experiment 1. 
Table~\ref{tab:OOS_filter_model} shows the accuracy of the proxy model, when we divide the annotated data into training and test sets with an 8:2 ratio, with a split of ten times. Note that the perfect accuracy in CB is because we identified only five OOS instances from our samples, which are extremely few. 

After applying LR or OOSF, we trained BERT models that serve the target task. For each dataset that applied LR without proxy models or used OOSF, we ran the training five times. For each dataset that used LR with proxy models, since each dataset from experiment 1 has been label-replaced five times, we ran training only once. With this approach, we acquired 15 model accuracy results for each task and condition.

\subsection{Results}

\subsubsection{Label Replacement}

\begin{figure}
    \centering
    \includegraphics[width=\linewidth]{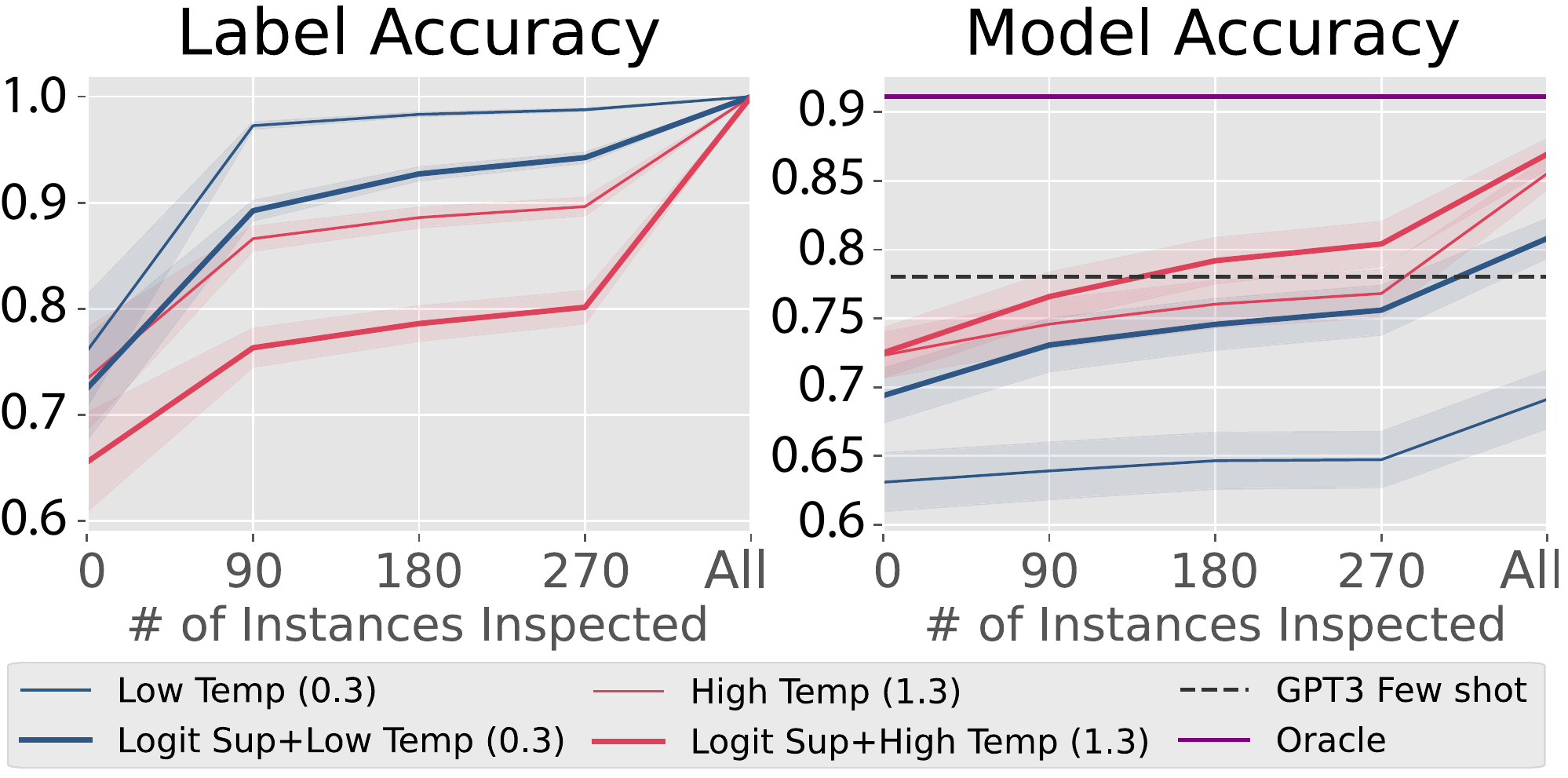}
    \caption{Impact of label replacement on label accuracy and model accuracy. Throughout this paper, error areas indicate 95\% confidence interval.}
    \label{fig:teaser}
\end{figure}

\begin{figure*}
    \centering
    \includegraphics[width=\textwidth]{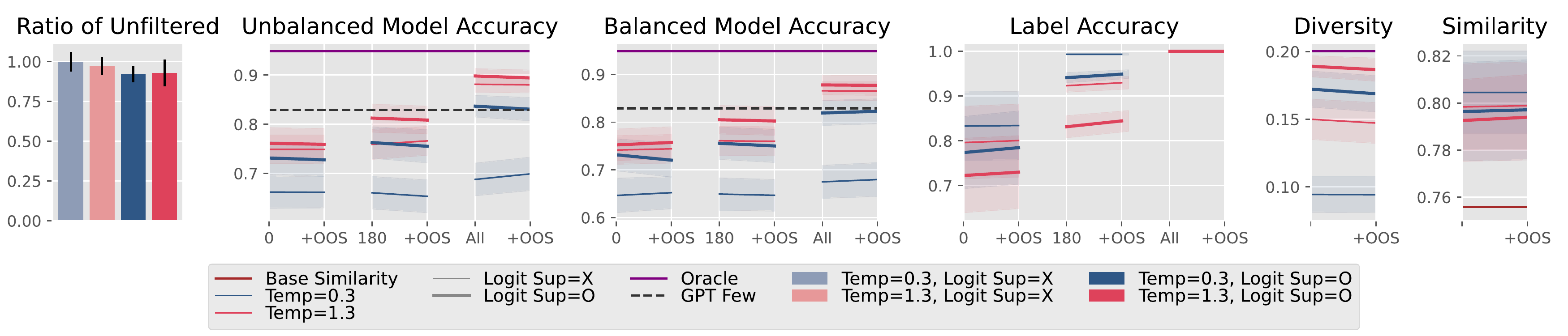}
    \caption{The ratio of instances filtered with OOSF, and its impact on model accuracy, label accuracy, diversity, and similarity, in aggregation across all tasks. As we examined the effect of OOSF with LR, for model accuracy and label accuracy, numbers left to +OOS indicate how many instances are inspected with LR.}
    \label{fig:exp3_whole}
\end{figure*}

Label Accuracy and Model Accuracy in Figure~\ref{fig:teaser} shows the results with LR. It shows how model accuracy and label accuracy change with the number of instances inspected (x-axis).
Other metrics, diversity, and similarity would not change with LR, as it keeps the texts as they are. 
For model accuracy, we also visualized the performance of oracle models and the GPT-3 few-/zero-shot classification. 

LR increases the model accuracy and label accuracy.
Moreover, with more labels inspected, the model accuracy and label accuracy further increased. 
LR also added more values to logit suppression. For example, without LR, using both high temperature (1.3) and logit suppression did not have a comparative benefit over using only high temperature. However, with label replacement, the addition of logit suppression started to benefit the model accuracy when using high temperature. When doing LR with proxy models, the benefit of logit suppression increased with more instances inspected, but with full LR, the size of this gap decreased a little bit.  
With LR of all instances, using both high temperature and logit suppression increased the absolute model accuracy by 17.8\%, compared to when using neither. It was greater than the increase from diversification approaches when LR was not used (9.4\%).
Furthermore, with high temperature and logit suppression, using LR on all instances could increase the absolute model accuracy by 14.4\% compared to not doing LR. 
When a high temperature and logit suppression are used together, the model accuracy outperformed GPT3's few-shot classification when LR was done for 180 instances. 
Across tasks, we found that specific patterns on how diversification approaches and LR impact the model accuracy can vary between tasks.
We provide details in Appendix~\ref{appendix:lr_detail_results}.

\subsubsection{Out-of-Scope Instances Filtering}

Figure~\ref{fig:exp3_whole} shows how many instances were filtered with OOSF and how it affects model accuracy, label accuracy, diversity, and similarity. 
We present model accuracy from both unbalanced and balanced data: when we balanced data, we used datasets with the same number of instances across different conditions by subsampling data with the smallest size of the filtered dataset. It was because filtering can make the number of instances different between conditions. 
For unbalanced data, we did not balance the number of instances. 

OOSF either increases or maintains label accuracy and similarity while decreasing or maintaining diversity, but there was no unified pattern of how they impact the model accuracy.
There tend to be few OOS-filtered instances without diversification approaches. 
For example, with a temperature of 0.3 and without logit suppression, OOSF removed very few data instances. 
Consequently, label accuracy, diversity, and similarity remained the same with OOSF. 
Without diversification approaches, the accuracy of trained models tends to be more unstable with large confidence intervals.
On the other hand, with diversification approaches, OOSF removed more instances, and hence there were slightly more changes in label accuracy, diversity, and similarity, with small increases in label accuracy and similarity while decreasing diversity.
However, in some cases, these changes were subtle or within the 95\% confidence intervals. 
Moreover, how the OOSF changes the model accuracy depends on the specific task and condition. 
We provide the OOSF results for each task in Appendix~\ref{appendix:oosf_detail_results}. 

\section{Conclusion}

In this work, we investigate approaches to harness LLMs and human efforts to generate text classification datasets with high accuracy and diversity. 
We study two text generation diversification approaches, 1) logit suppression, which restrains generating already frequently generated tokens, and 2) high temperature, which flattens the sampling probability of tokens. We found that they diversify text generation but hurt the accuracy in aligning specified labels with the generated data. 
We experiment with two human intervention approaches, 1) replacing misaligned labels with more adequate ones, and 2) filtering out-of-scope instances. 
We found that replacing labels makes diversification approaches more beneficial by increasing the accuracy of models trained with the generated dataset. 
On the other hand, efficient filtering of out-of-scope instances did not have a positive impact on the model accuracy. 
\section{Limitations}

Our implementation of proxy models applies those models after the whole data is generated.
Due to this, in the resulting dataset, the number of instances can often be unbalanced between labels. 
Such a limitation might be addressable by training proxy models from intermediate datasets with a smaller number of instances, and using those models while generating the rest of the dataset. As the data become unbalanced during the generation, the generation pipeline can try to generate more instances with labels that are a minority in the intermediate dataset.  
However, when we piloted this approach, we identified potential problems. 
First, intermediately trained proxy models could perform worse than those trained after all data are generated, due to the lower diversity in intermediate data used to train proxy models.
Second, if many data points generated with a specific label (label a) actually belong to another label (label b), there can be cases where most instances of label b come from the prompt with label a.
It can skew the linguistic patterns of instances within the dataset, as only a small number of texts for label b might have been from the prompt with label b. Advanced approaches to address these issues can be future work directions.

Our implementation of efficient OOSF was not effective in increasing model accuracy.
It might be due to the negative impact of removing instances, such as filtering instances on the decision boundary. 
As our study of OOSF was not complete, future work is necessary. 
Applying OOSF to the entire generated dataset and seeing the impact of their removal would be the first step. 
With a comprehensible understanding of OOSF, we would be able to design better OOSF strategies, such as filtering instances with various criteria. 

In this work, we only examined the \texttt{text-davinci-002} model of GPT-3. Although we believe that the overall trends of results would be similar for other models, examining other models with our approaches is a necessary future work. We also examined only one prompt (Prompt~\ref{prompt_A}), while there may be other options. In Appendix~\ref{sec:appendix_prompt_b}, we present partial results on using another prompt, showing that our approach is generalizable to other prompts. Combining human interventions with automatic annotation error detection~\cite{klie2023annotation} can be another future direction.

\section{Ethics Statement}
LLM-generated text data could have replicated biases within the used LLM. 
Diversification might alleviate such issues, as it steers the LLM to generate texts that it considers less probable, but bias can still exist after using the approach.
More human intervention approaches can be a potential solution. 
For example, the model builder can provide more specific prompts and examples to counter the biased generation~\cite{hartvigsen2022toxigen}. 
However, these approaches still would have limitations and how these approaches would impact the data bias and the resulting model performance would need to be further researched. 

\section*{Acknowledgements}
We want to thank Microsoft Research for supporting the work.

\bibliography{acl2023}
\bibliographystyle{acl_natbib}

\appendix

\begin{table*}[h]

\small
\centering
\begin{tabular}{p{0.06\textwidth}|p{0.15\textwidth}|p{0.7\textwidth}}
\noalign{\global\arrayrulewidth=0.5mm}
\hline
\noalign{\global\arrayrulewidth=0.15mm}
Task   & Text type                      & Label $\rightarrow$ Label in prompts \\
\noalign{\global\arrayrulewidth=0.5mm}
\hline
\noalign{\global\arrayrulewidth=0.15mm}
CARER  & emotional tweet                & joy $\rightarrow$ expressing joy,
                                          anger $\rightarrow$ expressing anger,
                                          fear $\rightarrow$ expressing fear, \newline
                                          sadness $\rightarrow$ expressing sadness, 
                                          love $\rightarrow$ expressing love,
                                          surprise $\rightarrow$ expressing surprise\\ \hline
CB     & news headline                  & non-clickbait $\rightarrow$ valid news,
                                          clickbait $\rightarrow$ clickbait \\ \hline
COLA   & sentence                       & grammatically acceptable $\rightarrow$ grammatically correct sentence, \newline
                                          grammatically unacceptable $\rightarrow$ grammatically incorrect sentence\\ \hline
FO     & sentence                       & informal $\rightarrow$ informal,
                                          formal $\rightarrow$ formal \\ \hline
HWU64  & human utterance to a chatbot & news $\rightarrow$ news, weather $\rightarrow$ weather, 
                                                            play $\rightarrow$ play, datetime $\rightarrow$ datetime, iot $\rightarrow$ iot, \newline
                                                            cooking $\rightarrow$ cooking, 
                                                            recommendation $\rightarrow$ recommendation, 
                                                            calendar $\rightarrow$ calendar, \newline
                                                            music $\rightarrow$ music,
                                                            takeaway $\rightarrow$ takeaway,  
                                                            lists $\rightarrow$ list, 
                                                            transport $\rightarrow$ transport, 
                                                            qa $\rightarrow$ qa, \newline
                                                            social $\rightarrow$ social, general $\rightarrow$ general, 
                                                            alarm $\rightarrow$ alarm, 
                                                            email $\rightarrow$ email, 
                                                            audio $\rightarrow$ audio
                                                            \\ \hline
PubMed & sentence from a medical paper  & objective $\rightarrow$ sentence about objective,
                                          methods $\rightarrow$ sentence about methods,
                                          results $\rightarrow$ sentence about results, 
                                          conclusions $\rightarrow$ sentence about conclusions, \newline
                                          background $\rightarrow$ sentence about background 
                                          \\ \hline
SST-2  & movie review          & positive $\rightarrow$ positive sentiment, negative $\rightarrow$ negative sentiment \\ \hline
SUBJ   & sentence from  a movie review   & objective $\rightarrow$ objective statement, subjective $\rightarrow$ subjective statement \\
\noalign{\global\arrayrulewidth=0.5mm}
\hline
\noalign{\global\arrayrulewidth=0.15mm}
\end{tabular}
\caption{Text types and labels used in prompts. \label{tab:text_type_and_label}}
\end{table*}

\section{Equation for Temperature Sampling}
\label{appendix:temp_samp}
Mathematically, with the temperature $T$ and original probability of token, $p_i$, the temperature sampled probability of token $i$, $f_{T}(p)_i$, would be denoted as below:

\begin{equation}
    f_{T}(p)_i = \frac{p_{i}^{1/T}}{\Sigma_{j}p^{1/T}_{j}}
\end{equation}

\section{Experiment 1 Details}
\subsection{Prompts Used in LLM Generation}
\label{sec:appendix_prompt_used}

\normalsize
For each task, we used prompt~\ref{prompt_A} with text types and labels as in Table~\ref{tab:text_type_and_label}. For example, for CB, a prompt can look like the below with examples:

\begin{equation}
\footnotesize
\parbox{\dimexpr\linewidth-3em}
{Write a \textbf{news headline} to cover all following elements\\
Elements: \textbf{valid news}\\
\textbf{News headline}: "Zach Johnson Wins Sony Open"

- - - - -

Write a \textbf{news headline} to cover all following elements\\
Elements: \textbf{clickbait}\\
\textbf{News headline}: "10 Of The Biggest Lies We Were Told In 2015"

- - - - -

Write a \textbf{news headline} to cover all following elements\\
Elements: \textbf{clickbait}\\
\textbf{News headline}:"
}
\tag{B}\label{prompt_B}
\end{equation}
\normalsize

\subsection{Sampling Oracle Dataset}
\label{appendix:oracle_dataset}

For the oracle dataset, if there are more than 5600 data points in the original dataset (CB, CARER, HATE, COLA, HWU64, SUBJ), we subsampled 5600 training data points. For SST2, we used all 6922 instances from the original dataset.
Note that these numbers are the same as the number of generated data instances.
For FO, we used the original training dataset as is (with 3622 data instances), as there are fewer than 5600 instances. 
For test datasets, from the same original dataset excluding instances used for the oracle dataset, we sampled 2400 data points for CB, CARER, HATE, and HWU64. For FO, COLA, SUBJ, and SST-2, we used the original test datasets as there were fewer than 2400 instances. 

\section{Results of the Experiment 1 on Individual Dataset}
\label{appendix:exp1_detail_results}
\begin{figure*}
    \centering
    \includegraphics[width=\textwidth]{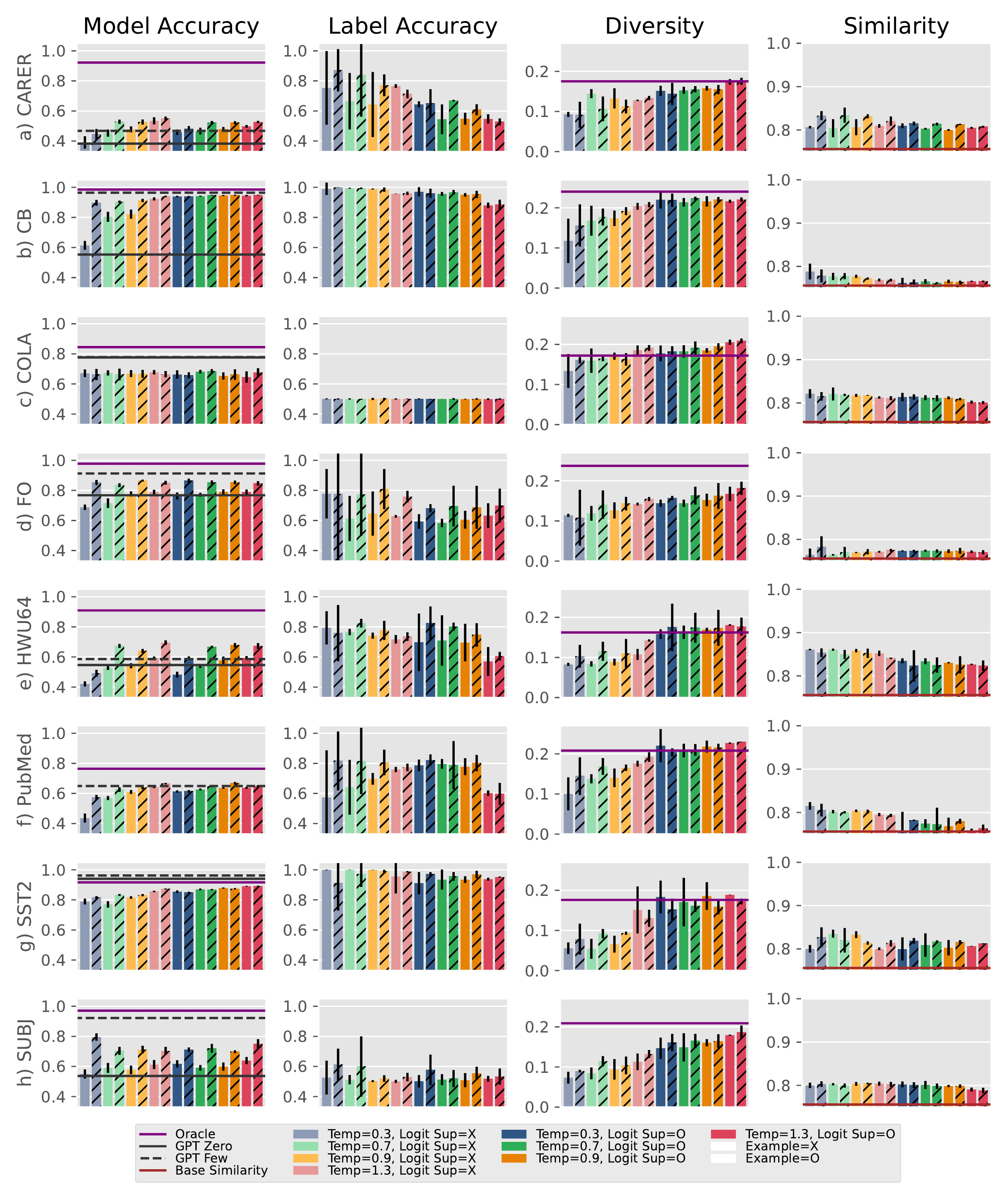}
    \caption{Impact of logit suppression and high temperatures on model accuracy, label accuracy, diversity, and similarity to the oracle dataset, for each task.}
    \label{fig:exp1}
\end{figure*}

Here, we introduce the result of the first experiment for individual tasks (Figure~\ref{fig:exp1}).

The benefit of logit suppression for each task depends on the combination of label accuracy, diversity, and similarity. Tasks that have high base label accuracy tend to improve model accuracy more with logit suppressions. For example, for CB and SST-2, those conditions with logit suppressions were clear winners in model accuracy over other combinations of approaches. For other tasks, where overall label accuracy tends to be lower, logit suppression did not have large benefits. COLA was the extreme case where the label accuracy was about 50\% in binary classification, indicating that the performance of the LLM in generating label-accurate instances was not better than random chance. In this case, logit suppression resulted in almost no increase in the model accuracy. Even in this case, logit suppression could increase the diversity of the generated text. 
With PubMed, we could observe an exception of label accuracy increasing with logit suppression when example seeding and high temperature (1.3) are not used (compare light and dark-colored unhatched bars in PubMed's Label Accuracy from Figure~\ref{fig:exp1}, except for red bars). It was because GPT-3 generates many similar errors without logit suppression and seeding examples. Specifically, without logit suppression, when prompted to write about the background sentence in a medical paper, GPT-3 generated many sentences starting with ``The purpose of this study was,'' which is more about the objective. 

\begin{table*}[t]
\small
\begin{tabular}{l|p{.44\textwidth}|l}
\hline
Task   & Example & Reason for filtering\\ \hline
CARER  & \texttt{No matter what life throws at you, always remember to find joy in the little things. \#HappyThoughts} & Not a self-expression of emotion              \\
CB     & \texttt{Valid News} & Not a news headline                 \\
SST-2  & \texttt{Jurassic World Fallen Kingdom} & Only movie title              \\
SUBJ   & \texttt{For what it's worth,} & Incomplete sentence and unable to decide subjectivity               \\ \hline
\end{tabular}
\caption{Examples of OOS instances.}
\label{tab:OOS_examples}
\end{table*}
\begin{figure*}
    \centering
    \includegraphics[width=.81\textwidth]{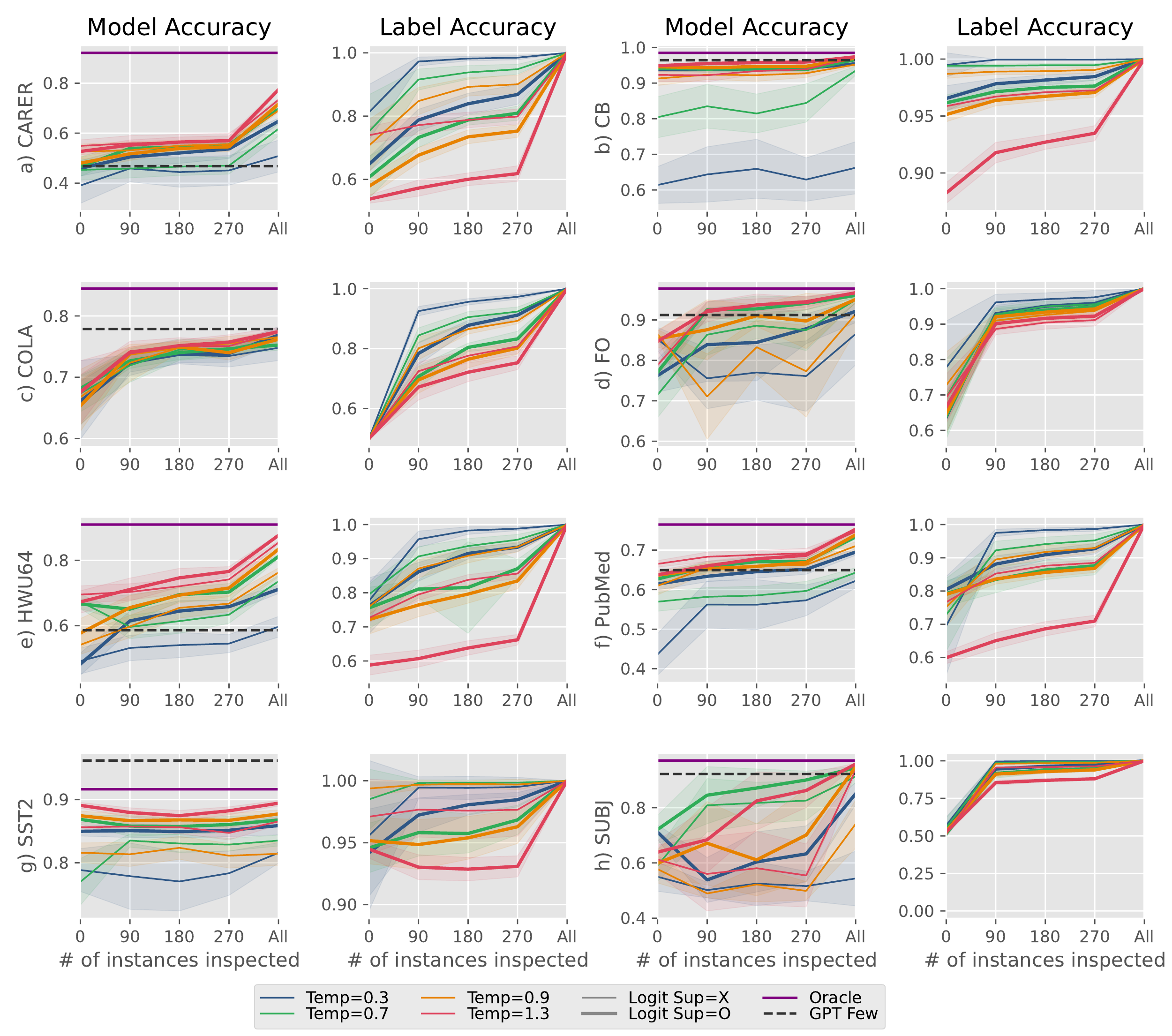}
    \caption{Impact of label replacement on model accuracy, label accuracy, for each task, on all temperature values.}
    \label{fig:exp2}
\end{figure*}
\begin{figure}
    \centering
    \includegraphics[width=0.5\textwidth]{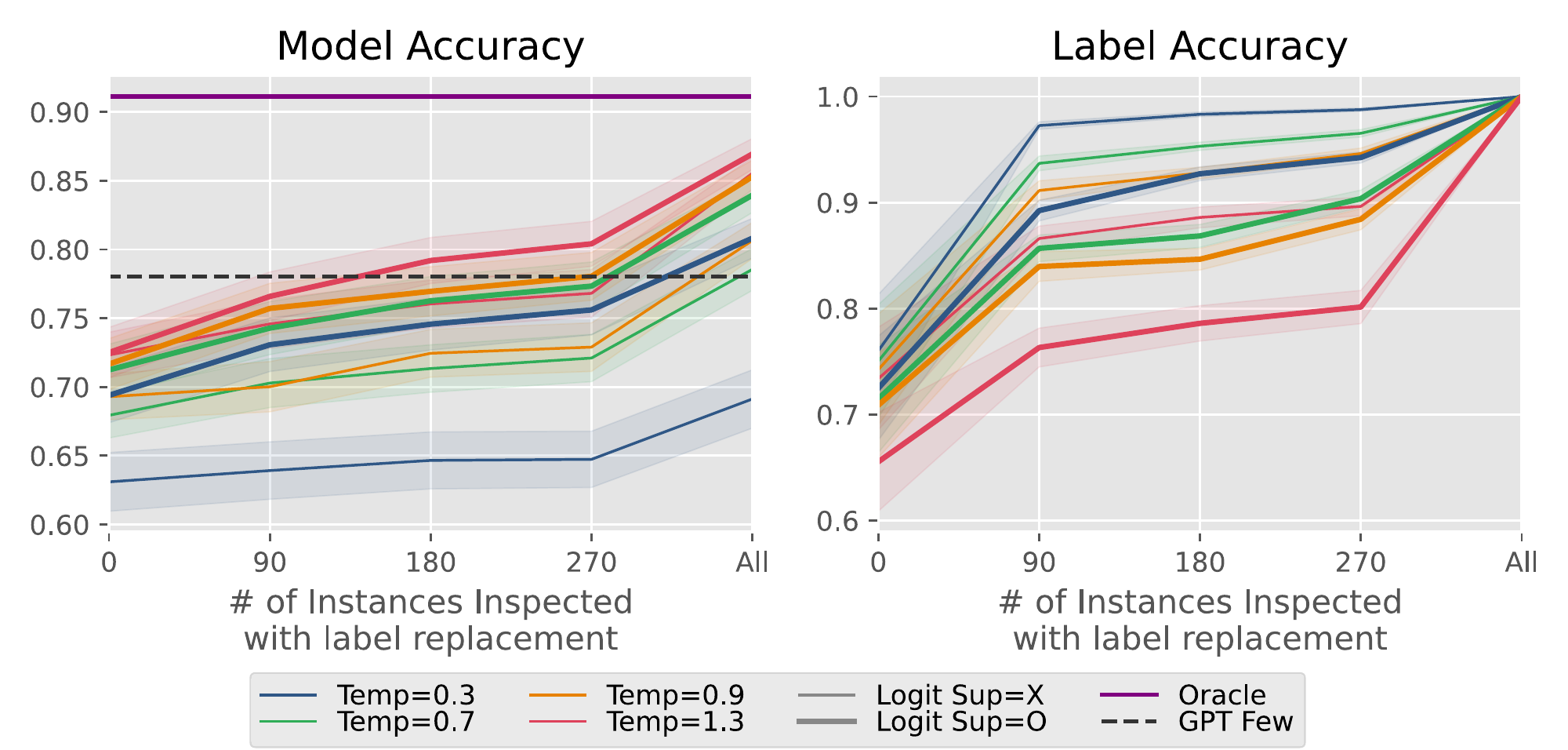}
    \caption{Impact of label replacement on model accuracy, label accuracy, for all tasks aggregated, on all temperature values.}
    \label{fig:LR_whole}
\end{figure}

For temperature also, specific patterns on how it affected label accuracy, diversity, and similarity differ between tasks.  
In PubMed, without logit suppression and example seeding, label accuracy even increased with higher temperatures, which was against the general pattern. 
In this case, similar to what we found with logit suppression, the lack of diversification approaches led to the generation of narrowly populated error instances. 
CARER was another case with the reversed trend: without logit suppression and seeding examples, the mean diversity was higher with a temperature of 0.7 than with a temperature of 1.3. 
It was because, with the high temperature of 1.3, many sentences started with ``I'm so,'' (on average 3012 occurrences) which was less the case for the lower temperatures of 0.7 and 0.9 (on average 841.5 occurrences).
In CARER, when example seeding and logit suppression are not used, label accuracy was also higher with the temperature of 1.3 than with lower temperatures, although the means were within 95\% confidence intervals. In this case, with lower temperatures of 0.7 and 0.9, more instances started with ``No matter what,'' which continues with advice on what to do in emotional situations. 
For such cases, no label is applicable since they are not the self-expression of emotions (on average, 32 occurrences with a temperature of 1.3 and 682.7 occurrences with temperatures of 0.7 or 0.9). 
Note that these are examples of out-of-scope instances.
Summarizing results of logit suppression and temperature sampling, these approaches increased diversity while hurting the label accuracy, but specific patterns could vary between tasks.

The utility of example seeding in label accuracy and model accuracy could also vary between tasks. For example, in the extreme case of COLA, examples did not increase label accuracy and model accuracy. How seeding examples impact the generation of data similar to the oracle dataset also depends on the task.  

\begin{figure*}
    \centering
    \includegraphics[width=.87\textwidth]{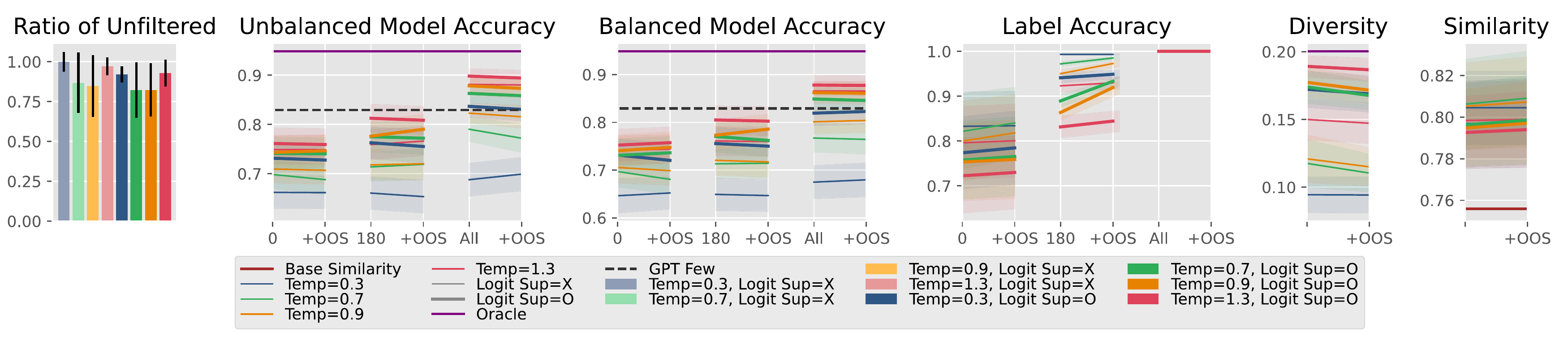}
    \caption{The ratio of instances filtered with OOSF, and its impact on model accuracy, label accuracy, diversity, and similarity, for all tasks aggregated, on all temperature values. As we examined the effect of OOSF with LR, for model accuracy and label accuracy, numbers left to +OOS indicate how many instances are inspected with LR.}
    \label{fig:exp3_temp_whole}
\end{figure*}

For CARER, HWU64, and PubMed in Figure~\ref{fig:exp1}, there were cases where the model accuracy was higher than the accuracy of GPT-3's few-shot learning. 
Other tasks showed lower accuracy than GPT-3's few-shot learning accuracy, indicating that GPT-3 few-shot classification can be a better alternative than training a model with generated data if the model builder has a budget to continuously access GPT-3 and is willing to hand over data through API. In Section~\ref{sec:exp2}, we show that human interventions can be a way to make the data generation approach applicable in more tasks by increasing the model accuracy higher than that of few-shot classifications from GPT-3. 

\section{Experiment 2 Details}
\subsection{Examples of OOS instances.}
\label{appendix:oos_instances}

\normalsize
We present examples of OOS instances in Table~\ref{tab:OOS_examples}.

\section{Results of the Experiment 2 on Varying Tasks}
\label{sec:appendix_temp}

We present the results of experiment 2 for individual tasks. Note that we also show results for all temperature values (0.3, 0.7, 0.9, and 1.3).

\subsection{Label Replacement}
\label{appendix:lr_detail_results}

Figure~\ref{fig:exp2} and~\ref{fig:LR_whole} shows the LR result for individual tasks and whole tasks aggregated, respectively, with all temperatures.
First, there were cases where logit suppression provided additional benefit upon high temperature only when LR was applied (comparing thick and thin red lines in Model Accuracy of CARER, HWU64, and PubMed in Figure~\ref{fig:exp2}).
Second, for tasks that already have high accuracy without LR (CB and SST-2), LR either resulted in very small model accuracy increases or even hurted the accuracy. 
For example, in SST-2, the label accuracy was already high without LR, and doing LR with proxy models could even decrease the label accuracy and model accuracy. 
Third, without diversification approaches, there were also cases where LR did not increase model accuracy much while label accuracy was greatly increased (thin blue lines in Model Accuracy of CARER, CB, FO, PubMed, SST2, SUBJ in Figure~\ref{fig:exp2}). It may show that fixing labels is more beneficial when there is enough diversity in the generated dataset. 
Fourth, CB, FO, and SUBJ were cases where models trained with generated data could outperform GPT-3's few-shot classification only with label replacement (some colored lines go over gray dashed lines with LR in Model Accuracy of CB, FO, and SUBJ in Figure~\ref{fig:exp2}). Among them, with FO, inspecting partial instances could also turn the model accuracy higher than that of GPT-3 few-shot classification. 
As expected, no approaches outperform oracle models as those models are used for LR. 
Fifth, for tasks with many classes (CARER, HWU64, and PubMed), when using LR with proxy models, the performance tends to increase not much dramatically as the number of annotated instances increases (Model Accuracy of CARER, HWU64, and PubMed in Figure~\ref{fig:exp2}).  
Higher model accuracy leaps occurred when all instances were inspected. 
It may indicate the difficulty of training accurate proxy models with many classes to consider.

\subsection{Out-of-Scope Filtering}
\label{appendix:oosf_detail_results}

\begin{figure*}
    \centering
    \includegraphics[width=.87\textwidth]{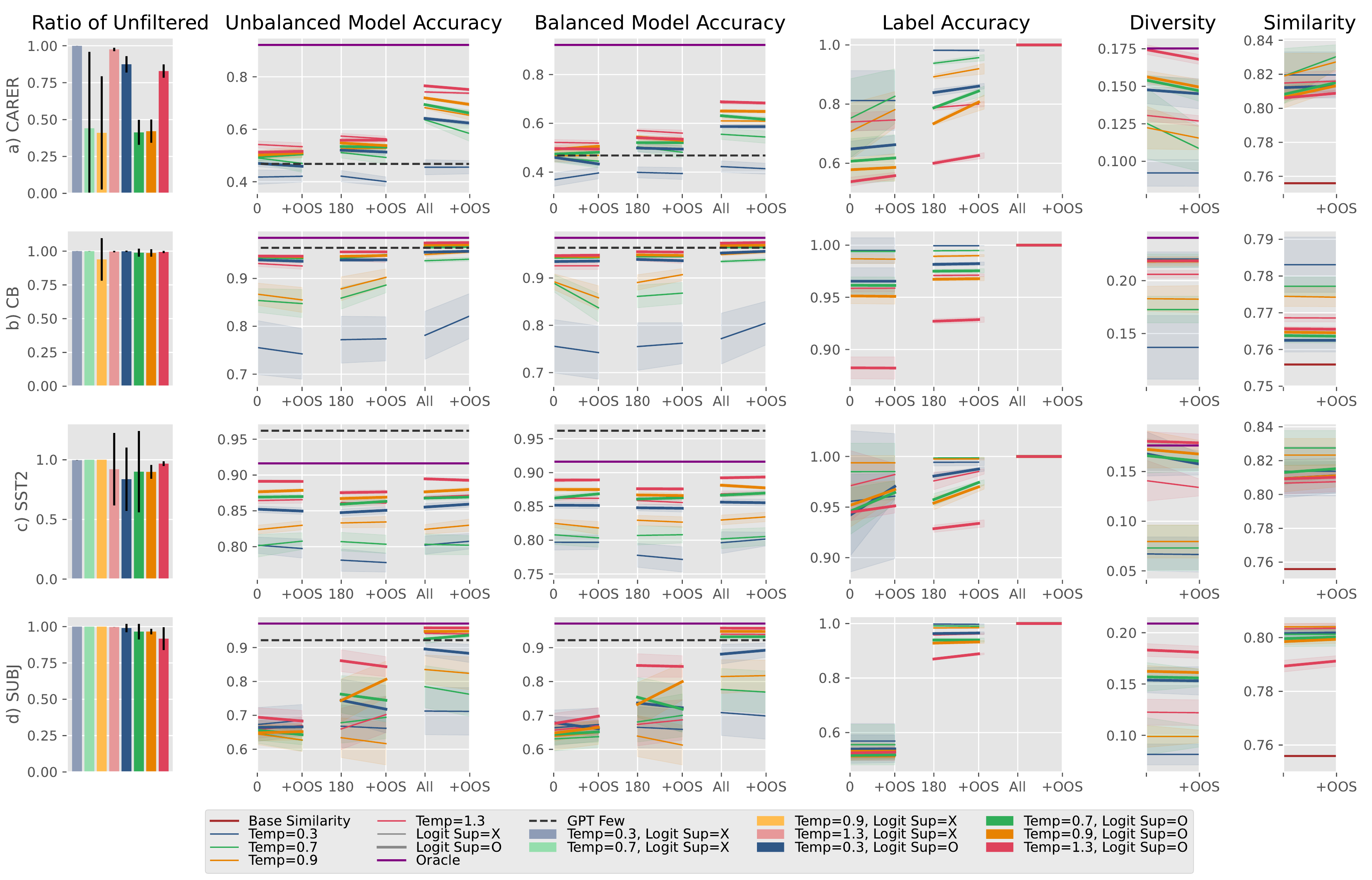}
    \caption{The ratio of instances filtered with OOSF, and its impact on model accuracy, label accuracy, diversity, and similarity, for each task, on all temperature values. As we examined the effect of OOSF with LR, for model accuracy and label accuracy, numbers left to +OOS indicate how many instances are inspected with LR.}
    \label{fig:exp3_tempall}
\end{figure*}

\begin{figure*}
    \centering
    \includegraphics[width=\textwidth]{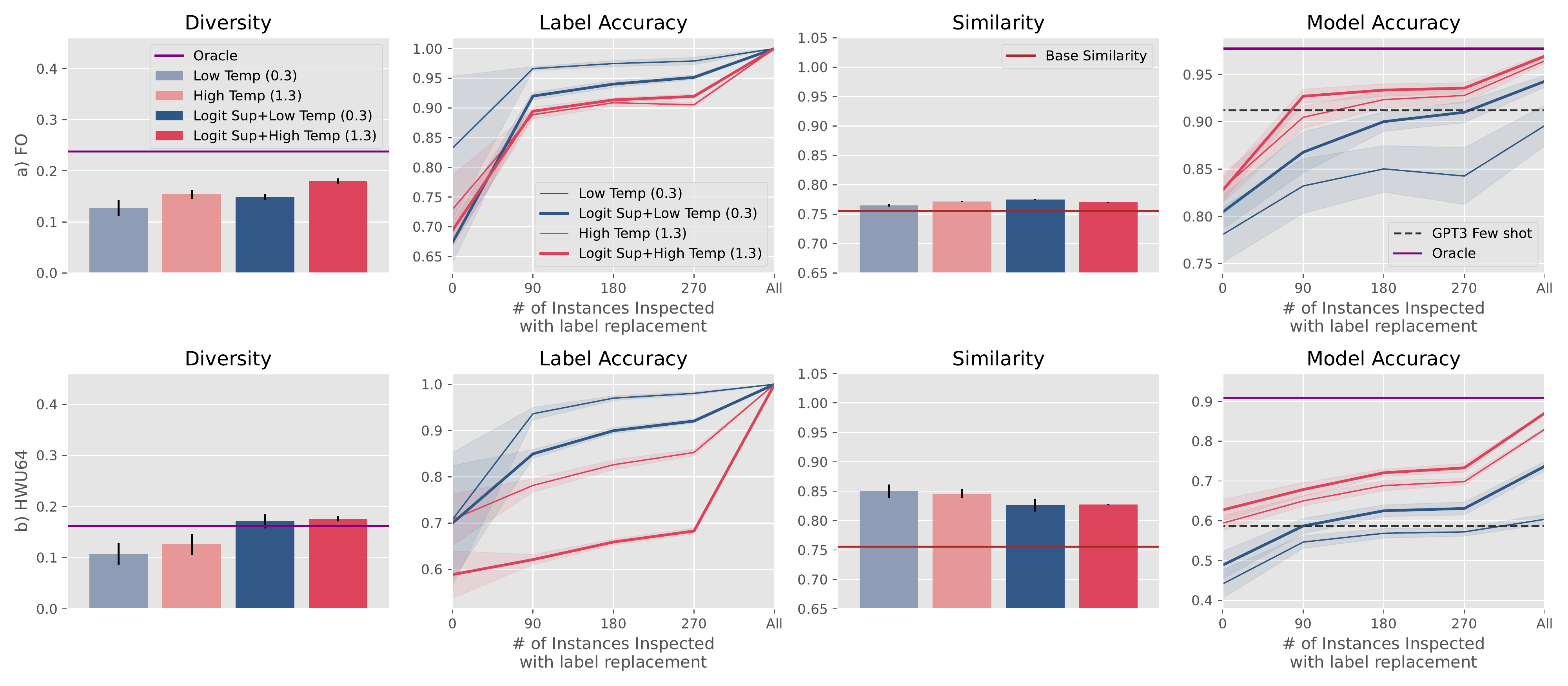}
    \caption{Result on prompt~\ref{prompt_C}.}
    \label{fig:exp_another}
\end{figure*}
 Figure~\ref{fig:exp3_temp_whole} and~\ref{fig:exp3_tempall} shows the OOSF results with all temperatures, for the aggregation of all tasks and individual tasks, respectively. As mentioned in the main text, it was difficult to find a general pattern of how OOSF impacts the model accuracy. Consistent patterns were that OOSF tends to increase or maintain label accuracy and similarity while decreasing or maintaining diversity. 

\section{Results on Prompt~\ref{prompt_C}}
\label{sec:appendix_prompt_b}

On two tasks (FO, HWU64), we conducted the experiment with another instructional prompt:
\begin{equation}
\footnotesize
\parbox{\dimexpr\linewidth-3em}
{Show me a \textbf{\texttt{text type}} that has the following characteristics

Characteristics: \textbf{\texttt{label}}

\textbf{\texttt{text type}}: "\texttt{Generated text}"
}
\tag{C}\label{prompt_C}
\end{equation}
\normalsize

We measured model accuracy, label accuracy, diversity, and similarity of generated datasets and also investigated how label replacement impacts label accuracy and model accuracy. The experiment setting was the same as the main experiment we conducted, except the prompt used. The trend in the results (Figure~\ref{fig:exp_another}) was similar to that of the prompt~\ref{prompt_A}.

\end{document}